\documentclass[10pt,twocolumn,letterpaper]{article}

\usepackage[pagenumbers]{cvpr} % To force page numbers, e.g. for an arXiv version

\usepackage[dvipsnames]{xcolor}

\definecolor{cvprblue}{rgb}{0.21,0.49,0.74}
\usepackage[pagebackref,breaklinks,colorlinks,citecolor=cvprblue]{hyperref}
\usepackage{multirow}
\usepackage{hhline}

\def\paperID{18} % *** Enter the Paper ID here
\def\confName{CVPR}
\def\confYear{2024}

\title{LEARN: A Unified Framework for Multi-Task Domain Adapt Few-Shot Learning}

\author{Bharadwaj Ravichandran, Alexander Lynch, Sarah Brockman, \\Brandon RichardWebster, Dawei Du, Anthony Hoogs, Christopher Funk\\
{\tt\small \{firstname.lastname\}@kitware.com}
\\Kitware Inc. %\\1712 Route 9, Suite 300, Clifton Park, NY 12065 USA
}

\begin{document}
\maketitle

%%%%%%%%% ABSTRACT
\begin{abstract}
     Both few-shot learning and domain adaptation sub-fields in Computer Vision have seen significant recent progress in terms of the availability of state-of-the-art algorithms and datasets.  Frameworks have been developed for each sub-field; however, building a common system or framework that combines both is something that has not been explored. As part of our research, we present the first unified framework that combines domain adaptation for the few-shot learning setting across 3 different tasks - image classification, object detection and video classification. Our framework is highly modular with the capability to support few-shot learning with/without the inclusion of domain adaptation depending on the algorithm. Furthermore, the most important configurable feature of our framework is the on-the-fly setup for incremental $n$-shot tasks with the optional capability to configure the system to scale to a traditional many-shot task. With more focus on Self-Supervised Learning (SSL) for current few-shot learning approaches, our system also supports multiple SSL pre-training configurations. To test our framework's capabilities, we provide benchmarks on a wide range of algorithms and datasets across different task and problem settings. The code is open source has been made publicly available here: \url{https://gitlab.kitware.com/darpa_learn/learn}
\end{abstract}

%%%%%%%%% BODY TEXT
\section{Introduction} \label{sec:Intro}
% Introduction split-up:
% Impact of few-shot learning and domain adaptation (40 -50$\%$)

% main contributions:
% incremental increase in number of labels from k shot to many shot
% disjoint class algorithm for domain adaptation and without. 
% first system to have unified framework for multiple types of problem with different inputs (image and video)
\begin{figure}[ht]
    \centering
    \includegraphics[width=1.05\linewidth]{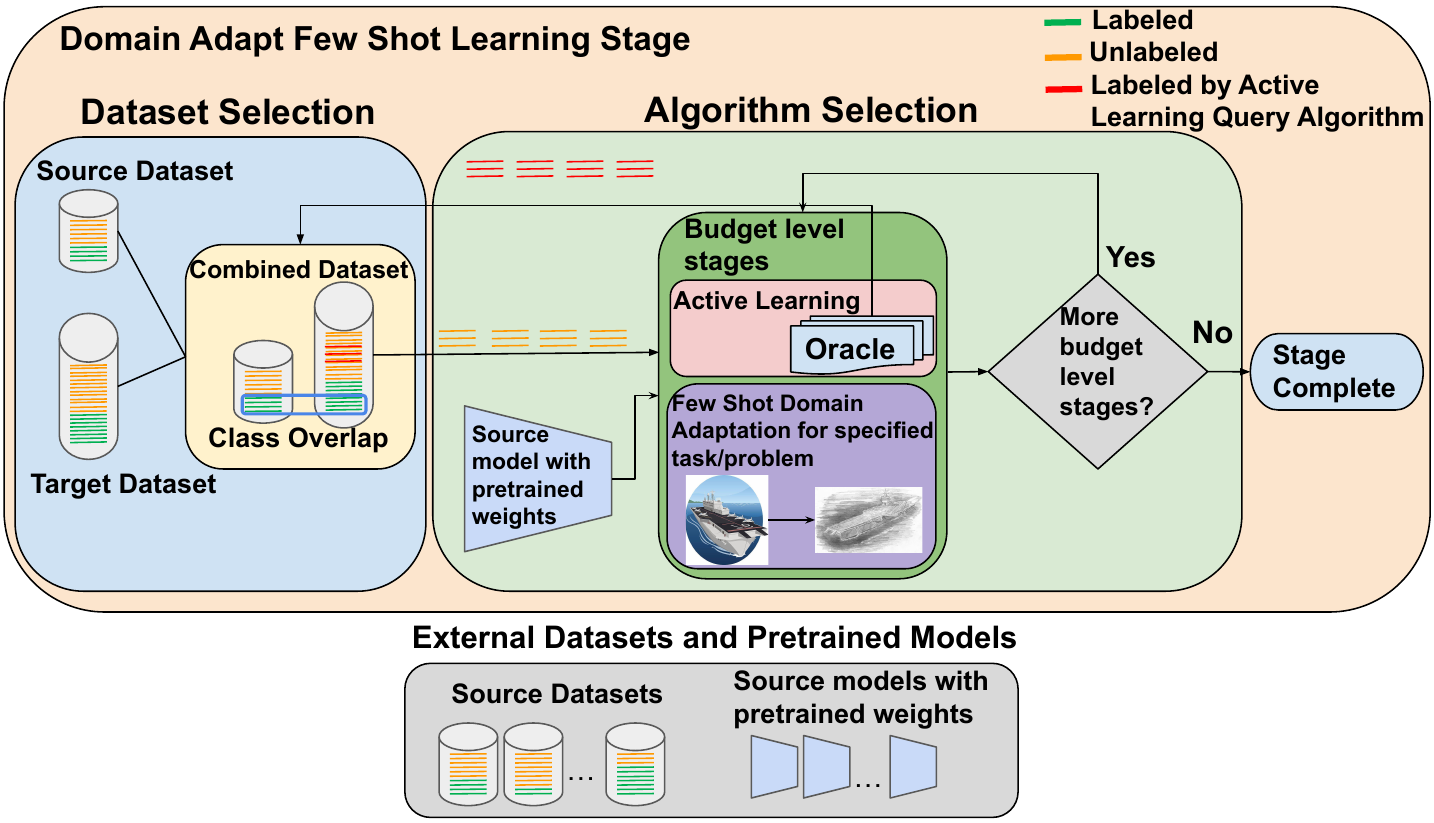}\vspace{-8pt}
    \caption{The LEARN Framework Concept Figure showing the different components of our Domain Adapt Few-Shot learning workflow. The Few-Shot domain adaptation block shows a sample images from DomainNet~\cite{peng2019moment} ClipArt and Sketch for the class ``Aircraft Carrier".
    \vspace{-15pt}
    }
    \label{fig:concept_figure}
    
\end{figure}
Up until 2018, with the advent of metric learning and meta-learning, the field of few-shot learning had little innovation with most methods revolving around pre-training and fine-tuning~\cite{song2022comprehensive}. However, after 2018, there has been an exponential increase in the number of few-shot learning publications with the majority of them coming within the field of computer vision. There have been a number of datasets adapted specifically for the task of few-shot learning such as mini-Imagenet~\cite{dhillon2020baseline}, CIFAR-10~\cite{krizhevsky2009learning}, and Meta-Dataset~\cite{triantafillou2020metadataset} which either subsample a larger dataset to restrict it to a few-shot setting or combine smaller datasets to increase the number of available classes. There has already been remarkable progress on these datasets with the accuracy on mini-Imagenet being pushed from $46.6\%$ in 2016 with Matching Nets~\cite{vinyals2017matching} to $95.3\%$ in 2022 with the P$>$M$>$F algorithm~\cite{hu2022pushing}. While image classification was the first computer vision task to see real attention using the few-shot paradigm, both object detection and video classification have also seen progress as both meta-learning and transfer learning have been applied with success~\cite{cao2020few,zhu2021closer,bar2022detreg}. 

The domain shift problem in computer vision is crucial when considering building systems around multiple datasets with varying image domains and class labels. Back in 2017, Tzeng \textit{et al.}~\cite{tzeng2017adversarial} proposed a simple-yet-efficient adversarial adaption baseline for domain transfer between the digit images of MNIST~\cite{lecun1998gradient} and USPS~\cite{291440}. Since then, the number of algorithms and datasets introduced as part of domain adaptation has seen a huge increase. For instance, in adversarial domain adaption, the MADA method~\cite{pei2018multi} introduced a fine-grained alignment using multiple domain discriminators (one for each class). Following this, many more methods like DADA~\cite{tang2020discriminative}, IDDA~\cite{kurmi2019looking}, RADA~\cite{wang2019adversarial} were introduced to push state-of-the-art performance across different datasets such as Office-31~\cite{saenko2010adapting} and Office-Home~\cite{venkateswara2017deep}. Unsupervised and Semi-Supervised Domain Adaption paved the way for a whole set of methods that uses Knowledge Distillation~\cite{liu2021graph,tang2021model,xiong2021source,yang2021transformer,yu2022source}, Statistical Domain Alignment~\cite{eastwood2021source,fan2022unsupervised,ishii2021sourcefree,zhang2021source}, Contrastive Learning~\cite{agarwal2022unsupervised, huang2021model,wang2022cross,xia2021adaptive}. In addition to the methods discussed above, a large number of public datasets have been made available for the Domain Adaptation Image Classification task~\cite{lecun1998gradient,291440,peng2019moment,venkateswara2017deep,saenko2010adapting}. Out of these, since the VisDA2019~\cite{visda2019} challenge, the DomainNet~\cite{peng2019moment} dataset is considered as a standard to advance state-of-the-art Domain Adaptation algorithms for fine-grained classification.
 
For object detection and video classification, there are a number of public datasets used in combination for domain adaptation. Cityscape~\cite{cordts2016cityscapes}, FoggyCityscape~\cite{Sakaridis_2018}, Sim10k~\cite{johnsonroberson2017driving}, and KITTI~\cite{liao2022kitti360} datasets are a group of overhead imagery datasets that are commonly used to train domain adaptation models using any two of the above mentioned datasets as the source and target.~\cite{oza2021unsupervised,yoo2022unsupervised}. PASCAL VOC is also commonly used as a source dataset which is then adapted to ClipArt1K~\cite{inoue2018crossdomain}, Watercolor2K~\cite{inoue2018crossdomain}, and BDD100K~\cite{yu2020bdd100k} for object detection domain adaptation~\cite{oza2021unsupervised}. For video classification, UCF101~\cite{soomro2012ucf101}, HMBD~\cite{6126543}, Kinetics~\cite{kay2017kinetics}, and ARID~\cite{xu2022arid} are used interchangeably for adapting from one dataset to another. These datasets are some of the oldest and most commonly used datasets for the video classification domain adaptation problem~\cite{Xu_2022,xu2022video,yang2022interact}. 

Therefore, with the increasing interest in few-shot learning and domain adaptation, there is a greater need for tools that can facilitate running few-shot and domain shift experiments. Existing few-shot learning frameworks like the LibFewShot~\cite{li2021LibFewShot} and $learn2learn$~\cite{Arnold2020-ss} provide a variety of APIs across different components of the framework pipeline, but are focused on only the image classification task. In this paper, we showcase a new framework which is easy to use across multiple computer vision tasks and provides APIs for algorithms to be applied in a domain-adapt few-shot setting. 

Furthermore, to provide a single cohesive workflow for training models in a variety of $n$-shot settings, we developed the iterative active learning workflow with an increasing number of label instances budgeted either on a per class basis or in totality as described in Figure \ref{fig:learn_process}. This modular training loop includes domain adaptation steps for at least one algorithm per task giving users the option to perform domain adaptation regardless of the task being executed. In order to balance the different algorithm and general framework parameters, we use a customizable configuration protocol in which the user specifies parameters to be used in their experiments either in default or through custom configuration files or as optional command line arguments for every possible parameter.

Based on our knowledge about existing domain-adapt few-shot learning frameworks, we believe our proposed system is the first unified framework to
\begin{itemize}[topsep=2pt, partopsep=0pt,itemsep=2pt,parsep=0pt]
    \item support multi-stage domain-adapt incremental $n$-shot learning, where $n$ is the number of labels per class; 
    \item support different computer vision tasks - image classification, object detection and video classification (with/without self-supervised pre-training based on the algorithm) - for different types of domain-adapt scenarios;
    \item provide a modular and scalable way to extend to a many-shot task without having to restart a few-shot experiment from scratch.
\end{itemize}
In order to support these claims, we conducted extensive experiments on the $3$ tasks mentioned above across different domain-adapt few-shot settings and have included the corresponding results in Section \ref{sec:evaluation}.
\section{Related Works} \label{sec:related_works}

Over the last few years, there has been significant development with respect to few-shot learning frameworks. Chada \textit{et al.}~\cite{chada-natarajan-2021-fewshotqa} created a NLP framework named ``FewshotQA" which leverages pre-trained text-to-text models for creating benchmarks for a Q$\&$A chatbot. Wang \textit{et al.}~\cite{DBLP:journals/tsp/WangBXZJ22} introduced a few-shot framework for 1D data, specifically Signal Modulation classification. 

In regard to vision-based few-shot learning frameworks, Lin \textit{et al.}~\cite{lin2022unified} introduced a unified framework for image classification and object detection that supports episodic learning methods, meta-learning and fine-tuning across multiple source/target datasets and pre-trained models. In addition to this, the authors propose a meta-dropout to improve the generalization capability during the meta-training stage. The benchmarks reported from a wide range of algorithms are primarily on $1,5$-shot tasks on the CUB~\cite{Wah2011TheCB} and mini-ImageNet~\cite{vinyals2016matching} datasets and $1,3,10$-shot tasks on the VOC2007~\cite{everingham2010pascal} dataset. LibFewShot is a similar framework from Li \textit{et al.}~\cite{li2021LibFewShot} which contains benchmarks across the different modes of few-shot learning, but focuses only on the image classification task. Furthermore, the framework supports $K$-way, $n$-shot task setups with benchmarks from different algorithms reported on the following datasets - mini-ImageNet~\cite{vinyals2016matching}, tieredImageNet~\cite{ren2018meta}, Stanford Dogs~\cite{khosla2011novel}, Stanford Cars~\cite{Krause_2013_ICCV_Workshops} and CUB~\cite{Wah2011TheCB}. Arnold \textit{et al.}~\cite{Arnold2020-ss} introduced the $learn2learn$ library which primarily focuses on Meta-learning few-shot benchmarks. The library consists of APIs to interact with PyTorch models and datasets and also contains high-level wrapper code for a bunch of existing meta-learning algorithms - MAML~\cite{finn2017model}, Meta-SGD~\cite{li2017meta}, MetaOptNet~\cite{lee2019meta} and ProtoNets~\cite{zhu2020robust}. \nocite{Kunchala_2023_WACV,Zhang_2022_WACV}
In contrast, our proposed framework is expanded to support three computer vision tasks in multi-stage domain-adapt few-shot learning.

Table \ref{tab:comparison} provides an overview of the functionalities of a couple of existing few-shot learning frameworks in comparison with the LEARN framework.

\begin{table}[t]
        \centering
        \resizebox{\columnwidth}{!}{\begin{tabular}{|c||c|c|c|}
        \hhline{|- - - -|}
        \textbf{Framework features} & \textbf{LEARN} & \textbf{LibFewShot \cite{li2021LibFewShot}} & \textbf{learn2learn \cite{Arnold2020-ss}} \\
        \hhline{|= = = =|}
        Customizable Datasets & \checkmark & \checkmark & \checkmark \\
        Customizable YAML Configuration file & \checkmark & \checkmark &  \\
        Domain Adaptation Capability & \checkmark & \checkmark & \\
        Iterative k-shot experiments & \checkmark & & \\
        Few-Shot Image Classification & \checkmark & \checkmark & \checkmark \\
        Few-Shot Video Classification & \checkmark & & \\
        Few-Shot Object Detection & \checkmark & & \\
        \hhline{|- - - -|}
        \end{tabular}}\vspace{-8pt}
        \caption{Summarization of the similarities between our framework and LibFewShot and learn2learn and the features we implemented to expand our framework's functionality and the ease of the user experience. \vspace{-15pt}}
        \label{tab:comparison}
\end{table}

\section{The LEARN Framework}\label{sec:framework}
The Label-Efficient Active Resilient Networks framework (LEARN) is a unified framework for multi-task, domain-adapt few-shot learning. % created as part of the DARPA Learning with Less Labeling (LwLL) program \cite{Lwll}. 
In the following subsections, we discuss the overall design and the experiment protocol configurations for the tasks and algorithms supported by the framework.

\begin{figure*}[t]
\centering
% \fbox{\rule{0pt}{2in} \rule{.9\linewidth}{0pt}}
\includegraphics[width=\linewidth]{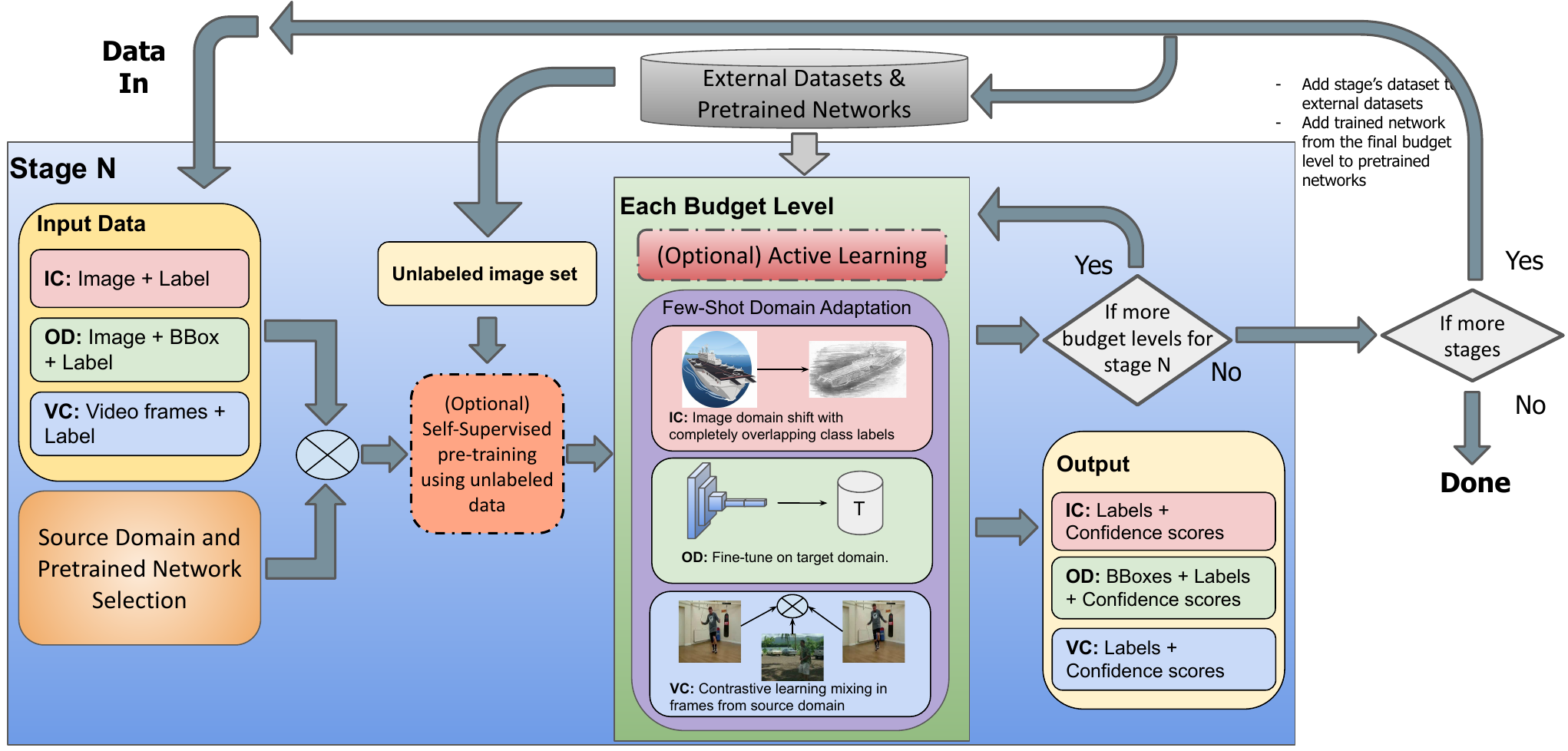}\vspace{-10pt}
\caption{Task workflow across the different tasks - \textbf{IC:} Image Classification, \textbf{OD:} Object Detection and \textbf{VC:} Video Classification from the LEARN framework. In addition to the multitask few-shot learning, the framework supports custom domain adapt scenarios as mentioned in the diagram along with the capability to setup SSL pre-training using unlabeled data. \vspace{-10pt}}
\label{fig:learn_process}
\end{figure*}

\subsection{Experiment Protocol}
Our framework is capable of running a variety of experiments for different tasks using different algorithms with a plethora of both general and algorithm specific hyperparameters. In order to give users access to the full breadth of experimental possibilities, the entire set of hyperparameters is exposed to the users to maximize ease of use. The experiment running protocol for the framework is configured with the help of Hydra~\cite{Hydra} which provides a customizable way to configure experiment parameters. A sample Hydra command looks for the following parameters to start an experiment - \texttt{problem.tasks\_to\_run}, \texttt{image\_classifier} (or) \texttt{video\_classifier} (or) \texttt{object\_detector} (based on the algorithm).
% (Comment, from Brandon) We should only show the parameters here that we're going to actually reference/discuss in the surrounding material.
These parameters point to individual JSON files containing the configuration details for the chosen task and algorithm respectively. Furthermore, selecting the source method pre-trained backbone weights is specified by \texttt{domain\_network\_selector.params.set\_source\\ \_data\_network}. One of the most important parameters is the \texttt{problem.add\_dataset\_to\_whitelist}. This provides a list of source datasets to consider during the protocol setup and skips all other datasets to save on processing time.

\subsubsection{Task Configuration}
There are many parameters to select between the task, algorithm, and experiment specific parameters. In order to make it easier for users to specify each of these parameters, the LEARN framework has a built-in interface which focuses on task JSON files that are defined locally. Each file contains the following task parameters:
\begin{enumerate}[topsep=2pt, partopsep=0pt,itemsep=2pt,parsep=0pt]
    \item \textbf{name:} A unique string identifier for the task.
    \item \textbf{problem\_type:} String value to denote 1 out of the 3 tasks supported by the framework - ``image\_classification", ``object\_detection" and  ``video\_classification".
    \item \textbf{stages:} A list of JSON objects (where each object corresponds to one stage) containing the following fields:
    \begin{enumerate}[topsep=2pt, partopsep=0pt,itemsep=2pt,parsep=0pt]
        \item \textbf{name:} Name of the domain-adapt stage - ``base" or ``adapt".
        \item \textbf{dataset:} Name of the target dataset for the stage.
        \item \textbf{seed\_budgets:} A list containing the cumulative $n$-shot budgets ($n$ labels per class), where each $n$-shot problem corresponds to the $(n-1)^{th}$ checkpoint during training and inference.
        \item \textbf{label\_budget:} A list containing the cumulative label budget across all classes (not $n$-shot). These additional labels are used during training after completing the $n$-shot training based off of the \texttt{seed\_budgets}.
    \end{enumerate}
    \item \textbf{whitelist:} A list containing the names of the source datasets that may be used for task. The dataset names not mentioned in the whitelist are skipped by the system during source dataset and method setup.
    \item \textbf{results\_file:} Name of the results files under the default ``outputs/$<$current\_date$>$/$<$experiment\_start\_time$>$" path that contains the predictions for each checkpoint of each stage.
\end{enumerate}

\subsubsection{Algorithm Setup}
As shown in Figure \ref{fig:learn_process}, each task has $N$ stages and a ``source" $\&$ ``target" dataset for each stage. Each stage consists of the following four algorithm-related sub-stages below that occur in an iterative manner based on the available label budget for each checkpoint. 
Furthermore, each of these sub-stages have their own dedicated sub-folders and are configured with default experiment parameters using YAML configuration files: 
\begin{enumerate}[topsep=2pt, partopsep=0pt,itemsep=2pt,parsep=0pt]
    \item \textbf{Domain/Network Selection:} Determine the source dataset and source pre-trained backbone based on the closest similarity to the chosen target dataset.
    \item \textbf{Algorithm Selection:} Choose the algorithm to run based on a specified task and given source $\&$ target dataset.
    \item \textbf{Active Learning Query Strategy:} Iteratively queries the unlabeled data that is most impactful for labeling based on the available label budget level (determined by \texttt{label\_budget}). By default, the querying algorithm does a random shuffle of the unlabeled target dataset and updates the training set based on the additional available label budget.
    \item \textbf{Few-Shot Domain Adaptation:} Perform the algorithm that creates a classification for the target task. After a particular budget level is reached, the adaptation stage performs an evaluation on the test data. 
\end{enumerate}
\section{Algorithms} \label{sec:algorithms}
In this section, we briefly describe the representative algorithms on three tasks used in our framework. The algorithms discussed below are chosen based on their availability for domain adaption scenarios (Image Classification) and self-supervised pre-training capability (for Object Detection and Video Classification) and thus supported by the LEARN framework.

\subsection{Image Classification}
\noindent\textbf{MetaBaseline}~\cite{chen2021meta} is a $2$-stage architecture for few-shot image classification. 
In the first classification stage, a classification model is trained using all the samples from base classes set and the last FC layer is removed to obtain the encoder. 
Following this, the average feature of per-class support set samples is computed and a given sample from a query set is classified using nearest-centroid with cosine similarity as the distance metric. 
In the second meta stage, the cosine similarity range is scaled using an additional learnable scalar before applying softmax during Meta-Baseline training. 
The goal of Meta-Baseline is to verify the efficacy of the meta-learning objective over a whole-classification model.

The benchmarking for Meta-Baseline includes a ResNet-12~\cite{he2016deep} backbone trained and evaluated on the miniImageNet~\cite{vinyals2016matching} and tieredImageNet~\cite{ren2018meta} few-shot classification datasets. In addition to these benchmarks, ResNet-18 and ResNet-50 backbones are trained and evaluated on the ImageNet-800~\cite{chen2021meta} dataset.

\noindent\textbf{MME} 
(MiniMax Entropy)~\cite{saito2019semi} was proposed to effectively align feature distributions of the source and target domains under a semi-supervised domain adaptation setting. The algorithm uses an adversarial optimization approach by updating the classifier's estimated class prototypes to maximize entropy on the unlabeled target domain and cluster features based on the estimated prototypes by minimizing entropy with respect to the feature extractor. 
As part of the maximization step, the entropy measured on unlabeled target samples shows the similarity between the estimated prototypes and target features. So, the weight vectors are modified to shift towards the target data by maximizing the entropy of the unlabeled target samples. Following this, as part of the minimization step, the feature extractor is updated to minimize entropy of the unlabeled target samples to enable better clustering around the estimated prototypes.

The proposed model consists of a feature extractor created by removing the last linear layer of an existing pre-trained model and adding a $K$-way linear classification layer with random initial weights. The feature extractors experimented by the authors include AlexNet, VGG16 and ResNet34 loaded with ImageNet pretrained weights and adapted for cross domain scenarios using the DomainNet~\cite{peng2019moment}, Office-Home~\cite{venkateswara2017deep} and Office-31~\cite{saenko2010adapting} datasets.

\noindent\textbf{PACMAC}
(Probing Attention-Conditioned Masking Consistency)~\cite{prabhu2022adapting} is a selection algorithm proposed for domain adaptation of self-supervised Vision Transformers (ViTs) to unseen domains. First, it performs in-domain self-supervised learning on a combination of source and target data to exploit task-discriminative features. Then, the algorithm selects reliable samples for self-training based on the predictive consistency across a set of attention-conditioned masks applied on a set of the target data.

During the attention-conditioned masking stage, a greedy assignment strategy is employed to select the high attention patches followed by generating its corresponding set of disjoint masks. After this, a consistency-based reliability metric is computed based on the original and masked images for selecting target samples for self-training.

Benchmark comparisons for the PACMAC system report target test set accuracy for three few-shot domain-adapt classification datasets - DomainNet~\cite{peng2019moment}, Office-Home~\cite{venkateswara2017deep} and VisDA2017~\cite{peng2017visda}. The model architecture consists of a ViT-Base model with 16x16 images patches, starting from two different pretrained backbones initialized with imagenet1k~\cite{russakovsky2015imagenet} weights - MAE~\cite{he2022masked} and DINO~\cite{caron2021emerging}.

\subsection{Video Classification}
\noindent\textbf{X-Clip}~\cite{ni2022expanding} is an expansion of CLIP~\cite{radford2021learning} specifically designed for the task of video recognition. The key contribution is a cross-frame 
attention mechanism for encoding long range temporal dependencies across frames. This transformer takes in raw frame data and generates frame level embeddings while allowing information to be shared across frames during the embedding process. A second ``multi-frame integration transformer"~\cite{ni2022expanding} is then used to fuse those frame-level embeddings and output a single embedding for the whole video.

The X-Clip algorithm then uses a learned text encoder to augment the raw label text embeddings. This text encoder first uses a multi-head attention layer to create a joint embedding of the encoded raw label text and the cohesive video embedding. This text-video embedding is then fed through a FFN (feed forward network) and outputs a text embedding based on the raw label that is specific to the context of the provided video. 

Finally a cosine similarity is calculated between the augmented label text embedding and the full video embedding. The goal of this training is to maximize the cosine similarity between these embeddings.

\noindent\textbf{TimeSformer}~\cite{bertasius2021spacetime} implements a transformer architecture that enables feature learning across the temporal and spatial dimensions. The algorithm takes as input a set of frames and then splits each frame into a set of \(N\) non-overlapping square patches. The patches are then mapped to an embedding vector through a learned embedding matrix, and these embeddings are then fed as inputs into the transformer.

The authors experiment with four different variations of self-attention blocks, but the one that achieves the highest success and the one we focus on is referred to as the ``Divided Space-Time Attention (T +S)"\cite{bertasius2021spacetime} block. In this block, temporal attention is calculated first by comparing each patch with the patch at the same location across all the given frames. The encoding from the temporal attention block is then fed back into the spatial attention block which compares all of the patches across a single frame. The encoding given from the spatial attention block is then passed through a final multi-layer perceptron (MLP) module to get the final encoding for a given patch. 

\noindent\textbf{CoMix}
(Contrast and Mix)~\cite{alaniz2022compositional} uses contrastive learning by maximizing the similarity between both the same video played at different playback speeds as well as different videos played back at varying playback speeds. 
%The algorithm starts by breaking up a given video into a ``fast" video composed of \(f\) clips and a ``slow" video composed of \(s\) clips where \(s < f\). 
Sets of fast ($f$) and slow ($s$) clips are used as inputs to a feature encoder which maps them to a set of feature encodings. 
A temporal graph encoder is then used to create a fully connected graph on top of the clip-level feature embeddings. The graph representation is then fed into a graph convolutional network which produces video classification output with per-class confidences.
% and finally the node features are put through an average pooling function to get a sequence of confidences the length of the number of video classification classes are being used.

% By running the same video at different speeds, positive pairs are created such that the cosine distance between the confidences should be minimized. Alternatively different videos at different speeds for negative pairs and should have the distances between them maximized. 
Another unique contribution of CoMix is the mixing of background frames in a video.
% Synthetic videos are created by mixing the background frames (frames in which no activity is occurring) from and using them to replace the background frames from a video from a separate domain. This synthetic video will then posses the action semantics of the original video but also possess frames from a separate domain.
Synthetic videos created by cross-domain background frame mixing has the action semantics of the original video but also consists of frames from a different domain. This allows additional positive examples to be created for any given video to be utilized in the contrastive learning process.

\subsection{Self-Supervised Object Detection Pre-training}
\noindent\textbf{DETReg}~\cite{bar2022detreg} While most object detection pre-training algorithms focus on just training the backbone in the pre-training step, DETReg~\cite{bar2022detreg} pre-trains both the embedding backbone as well as the localization head by carrying out an object localization task as well as an object embedding task. DETReg's object localization pre-training ``uses simple region proposal methods for class-agnostic bounding-box supervision" by using the Selective Search~\cite{selectivesearch} method. Selective search uses various visual cues to generate a set of bounding boxes around the predicted objects in the image excluding any class predictions. The object localization task takes a set of boxes output by the selective search unsupervised region proposal network (RPN) and optimizes a loss that minimizes the difference between the boxes output by the detector and the boxes generated from the RPN.

When pre-training the object embedding task, the encoders learn transformation-invariant embeddings so that the detectors trained are robust to variation in image transformations such as cropping or translation. To accomplish this, a pre-trained SwAV~\cite{caron2021unsupervised} is used to generate a ground truth feature embedding to train the object detection classification head.  

\noindent\textbf{CutLER} (Cut and Learn)~\cite{wang2023cut} is agnostic to the underlying detector being used. It is comprised of two sections that are repeated for multiple rounds. First CutLER uses a novel expansion of the NCut~\cite{ncut} algorithm, referred to as MaskCut, that iteratively runs through the NCut algorithm detecting \(x\) objects where \(x\) is $3$ by default. The MaskCut algorithm produces a series of binary masks for each object detected. 
After generating the set of binary masks, a loss function called DropLoss is applied. DropLoss is specifically designed to encourage the discovery of the ``ground-truth" objects that were not discovered by the initial MaskCut algorithm. It accomplishes this by dropping the loss for each predicted region that has an overlap greater than or equal to a set threshold. 

By using this coarse mask generation from MaskCut and the DropLoss function that encourages new objects to be detected, multiple rounds of this self-training process is applied to steadily increase the number of ``ground-truth samples" obtained from a given image.
\section{Datasets}
The datasets discussed below are commonly used public datasets that vary in terms of the granularity of the classes and also provide the domain adaptation scenarios (for Image Classification) that could be used to evaluate our framework's capabilities.

\begin{figure}[t]
    \centering
    \includegraphics[width=0.5\textwidth]{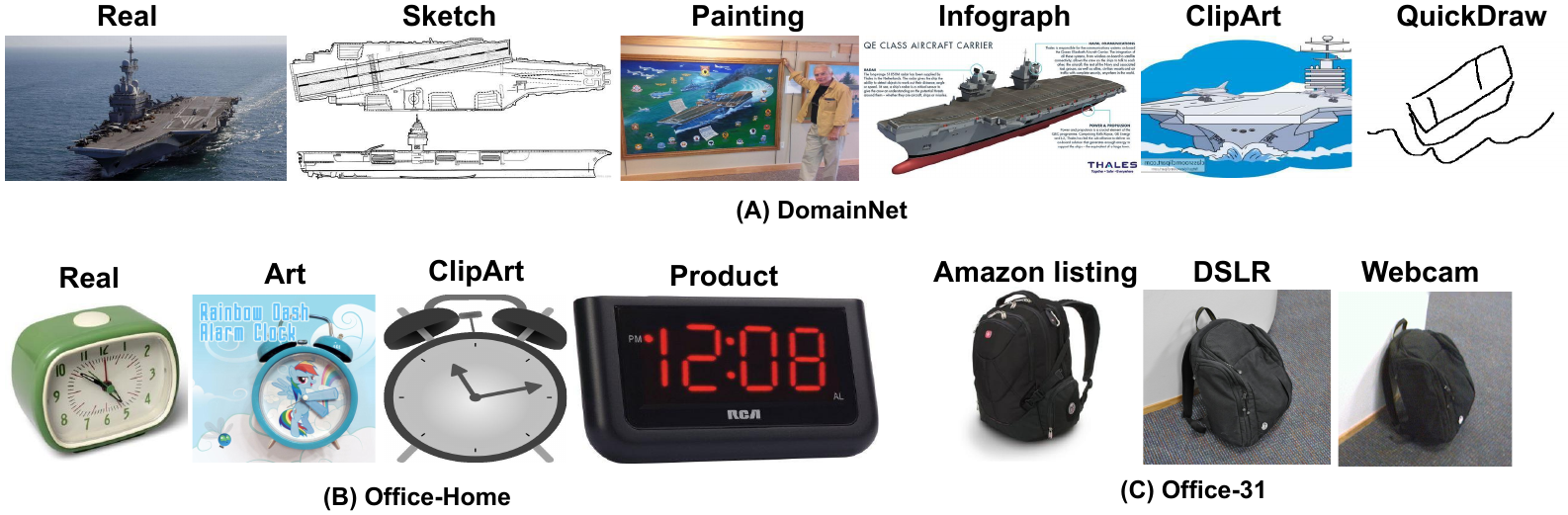} \vspace{-15pt}
    \caption{Overview of the image domains across the $3$ image classification datasets used in our benchmarking. \textbf{A}: DomainNet dataset samples for the class ``Aircraft Carrier". \textbf{B}: Office-Home dataset samples for the class ``Alarm Clock". \textbf{C}: Office-31 dataset samples for the class ``Backpack".} \vspace{-10pt}
    \label{fig:img_class_datasets}
\end{figure}

\begin{figure}[t]
    \centering
    \includegraphics[width=0.5\textwidth]{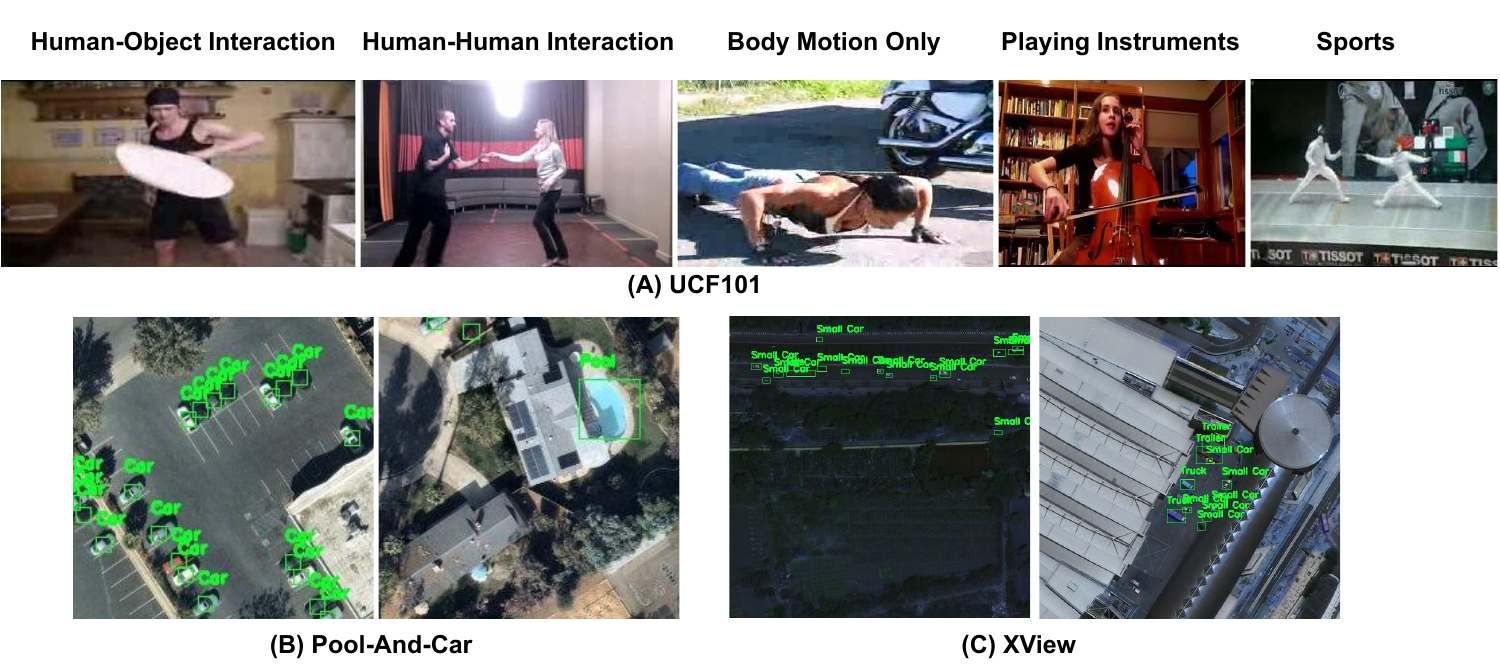} \vspace{-15pt}
    \caption{Overview of the Video Classification (A) and Object Detection (B and C) datasets used in our benchmarking comparisons. \textbf{A}: UCF101 with its $5$ major categories. \textbf{B}: Pool-And-Car sample images overlayed with pool and car detections. \textbf{C}: XView sample images overlayed with detections of ``Small car", ``Truck" and ``Trailer" classes.} \vspace{-10pt}
    \label{fig:vc_od_datasets}
\end{figure}

\noindent\textbf{DomainNet}
\cite{peng2019moment} consists of about 600k images with $345$ overlapping categories across $6$ different image domains: \textit{Real, Sketch, ClipArt, Quickdraw, Painting}, and \textit{Infograph}. Figure \ref{fig:img_class_datasets}(A) shows image samples for the ``Aircraft Carrier" class across the different image domains. For our experiments, we consider $3$ out of the $6$ domains (Real, Sketch, ClipArt) with all classes included.

\noindent\textbf{Office-Home}
\cite{venkateswara2017deep} consists of about 15.5k images with $65$ overlapping categories across $4$ domains. Figure \ref{fig:img_class_datasets} (B) contains examples for the ``Alarm Clock" class from the different image domains. For our benchmark comparisons, we will be using the image samples from the Real, Art and ClipArt domains.

\noindent\textbf{Office-31}
\cite{saenko2010adapting} consists of totally 4.1k images with $31$ overlapping categories across $3$ domains. Figure \ref{fig:img_class_datasets} (C) contains examples for the ``Backpack" class from 3 domains - Amazon, DSLR, and Webcam.

\noindent\textbf{PoolCar}
\cite{poolcar} is made up of overhead imagery with $2$ classes labeled (Pool and Car). The dataset is split into a training set with $2,998$ images and a test set with $750$ images. Due to the presence of only two classes, the regression task of predicting the bounding boxes is much more challenging than the classification task of identifying the category of an object.

\noindent\textbf{xView}
\cite{lam2018xview} contains over $1$ million individual object instances labeled as one of $60$ classes. The sizes of these labeled object vary highly ranging from objects $10$ pixels wide to $10,000$ pixels wide. Additionally xView provides a granularity of classes as $80\%$ of the classes are specific sub-classes of a given parent class, \textit{e.g.,} ``Pickup Truck'', ``Utility Truck''.

\noindent\textbf{UCF101}
\cite{soomro2012ucf101} contains over $27$ hours of video with over 13k clips annotated as one of $101$ classes. These $101$ classes can be broken down into $5$ major categories that are shown in Figure \ref{fig:vc_od_datasets}. Released in 2012, it is one of the most bench marked activity classification datasets and contains all public data.
\section{Evaluation}\label{sec:evaluation}
\subsection{Framework and Experiment Setup}
The LEARN framework is structured as specified in Section~\ref{sec:framework}. Before that, to install the framework, there is a README that goes through a series of python conda and pip installations that sets up the dependencies based on a ``requirements.txt" file and installs the general dependencies of the framework. In addition to this, each algorithm folder has specific dependencies installed as part of the setup process.

With the help of the configurations discussed in Section~\ref{sec:framework}, we have a straightforward way of setting up the different task and algorithm parameters needed for a specific experiment. Furthermore, using the Hydra command, we can override a wide range of experiment parameters during runtime. 

\subsection{Result Analysis}
\begin{figure}[t]
    \centering
    \includegraphics[width=0.5\textwidth]{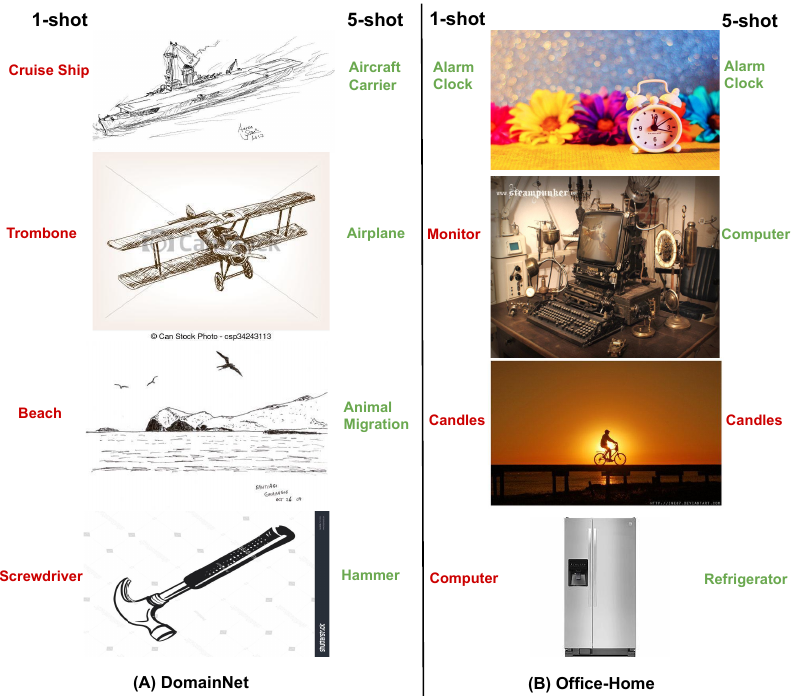} \vspace{-18pt}
    \caption{Qualitative results for the Few-Shot Domain Adapt Image Classification task using the PACMAC~\cite{prabhu2022adapting} algorithm on the (A) DomainNet Sketch (B) Office-Home Art datasets. \vspace{-10pt}}
    \label{fig:img_cls_qual_results}
\end{figure}

\begin{figure}[t]
    \centering
    \includegraphics[width=0.5\textwidth]{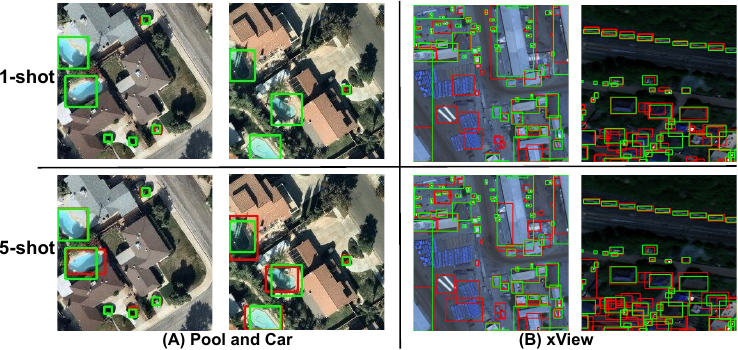} \vspace{-18pt}
    \caption{Qualitative results (\textcolor{green}{GT}, \textcolor{red}{Detections}) for the Self-Supervised Pretraining Object Detection task using the CutLER~\cite{wang2023cut} algorithm on the (A) Pool and Car (B) xView datasets. \vspace{-20pt}}
    \label{fig:obj_det_qual_results}
\end{figure}

\begin{figure}[t]
    \centering
    \includegraphics[width=0.5\textwidth]{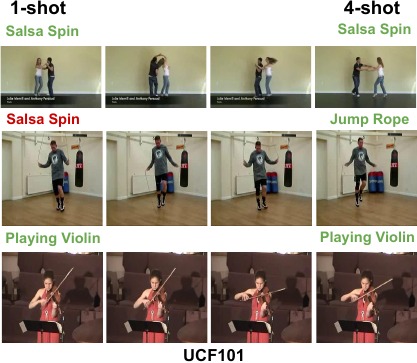} \vspace{-15pt}
    \caption{Qualitative results for Video Classification task using the TimeSformer~\cite{bertasius2021spacetime} algorithm on the UCF101 dataset. \vspace{-15pt}}
    \label{fig:vid_cls_qual_results}
\end{figure}

\begin{table*}[t]
        \resizebox{0.625\textwidth}{!}{\begin{tabular*}{\textwidth}{|c|c|c|c|c|c|c|c|c|c|c|c|c|c|}
            \cline{1-14}
            \multirow{2}{*}{Algorithm} & \multicolumn{2}{|c|}{Pretraining Backbone} & \multicolumn{3}{|c|}{Dataset} & \multicolumn{4}{|c|}{Seed budgets ($n$-shot)} & \multicolumn{4}{|c|}{Label budget (total labels $N$)} \\
            \cline{2-14}
            & Model & Weights & Source & Target\_Base & Target\_Adapt & 1-shot & 2-shot & 5-shot & 10-shot & \text{\footnotesize$\sim$0.1$N$} & \text{\footnotesize$\sim$0.25$N$} & \text{\footnotesize$\sim$0.5$N$} & \text{\footnotesize$N$} \\
            \hhline{|= = = = = = = = = = = = = =|}
            \multicolumn{14}{|c|}{Image Classification} \\
            \hhline{|= = = = = = = = = = = = = =|}
            \multirow{3}{*}{MetaBaseline~\cite{chen2021meta}} & ResNet18 & ImageNet1k & ImageNet1k & DomainNet-ClipArt & N/A & 0.107 & 0.155 & 0.274 & 0.328 & 0.328 & 0.356 & 0.370 & 0.379 \\
            & ResNet18 & ImageNet1k & ImageNet1k & OfficeHome-ClipArt & N/A & 0.211 & 0.323 & 0.452 & 0.513 & 0.513 & 0.526 & 0.537 & 0.545 \\
            & ResNet18 & ImageNet1k & ImageNet1k & Office31-Webcam & N/A & 0.654 & 0.792 & 0.874 & 0.887 & N/A & N/A & N/A & 0.881 \\
            \cline{1-14}
            \multirow{3}{*}{MME~\cite{saito2019semi}} & EfficientNet-B2 & ImageNet1k & DomainNet-Real & DomainNet-ClipArt & DomainNet-Sketch & 0.466 & 0.478 & 0.518 & 0.558 & 0.559 & 0.599 & 0.600 & 0.598 \\
            & EfficientNet-B2 & ImageNet1k & OfficeHome-Real & OfficeHome-ClipArt & OfficeHome-Art & \textbf{0.638} & \textbf{0.650} & \textbf{0.670} & \textbf{0.689} & 0.673 & 0.677 & 0.677 & 0.666 \\
            & EfficientNet-B2 & ImageNet1k & Office31-Amazon & Office31-Webcam & Office31-DSLR & 0.980 & 0.950 & 0.960 & 1.0 & N/A & N/A & N/A & 0.990 \\
            \cline{1-14}
            \multirow{3}{*}{PACMAC~\cite{prabhu2022adapting}} & EfficientNet-B2 & ImageNet1k & DomainNet-Real & DomainNet-ClipArt & DomainNet-Sketch & \textbf{0.488} & \textbf{0.510} & \textbf{0.544} & \textbf{0.566} & N/A & N/A & N/A & N/A \\
            & EfficientNet-B2 & ImageNet1k & OfficeHome-Real & OfficeHome-ClipArt & OfficeHome-Art & 0.494 & 0.547 & 0.576 & 0.609 & N/A & N/A & N/A & N/A \\
            & EfficientNet-B2 & ImageNet1k & Office31-Amazon & Office31-Webcam & Office31-DSLR & \textbf{1.0} & \textbf{1.0} & \textbf{1.0} & \textbf{1.0} & N/A & N/A & N/A & N/A \\
            \cline{1-14}
            \hhline{|= = = = = = = = = = = = = =|}
            \multicolumn{14}{|c|}{Object Detection} \\
            \hhline{|= = = = = = = = = = = = = =|}
            DETReg~\cite{bar2022detreg} & Convnext FC-2048 & Coco 2014 & Coco 2014 & Pool and Car & N/A & 0.22 & 0.44 & 0.63 & 0.66 & \textbf{0.75} & 0.75 & \textbf{0.84} & \textbf{0.90}\\
            CutLER~\cite{wang2023cut} & Convnext FC-2048 & Coco 2014 & Coco 2014 & Pool and Car & N/A  & \textbf{0.23} & \textbf{0.49} & \textbf{0.64} & \textbf{0.71} & 0.72 & \textbf{0.77} & 0.83 & 0.89  \\
            \hhline{|= = = = = = = = = = = = = =|}
            \multicolumn{14}{|c|}{Video Classification} \\
            \hhline{|= = = = = = = = = = = = = =|}
            X-Clip~\cite{ni2022expanding} & & & Kinetics400 & ucf101 & N/A & \textbf{0.843} & 0.889 & 0.950 & 0.969 & 0.964 & 0.976 & 0.978 & 0.977\\
            TimeSformer~\cite{bertasius2021spacetime} & & Kinetics400 & Kinetics400 & ucf101 &  N/A & 0.795 & \textbf{0.897} & \textbf{0.957} & \textbf{0.981} & \textbf{0.982} & \textbf{0.990} & \textbf{0.990} & \textbf{0.991} \\
            CoMix~\cite{alaniz2022compositional} & & & Kinetics400 & ucf101 & N/A & 0.609 & 0.806 & 0.949 & 0.968 & 0.966 & 0.968 & 0.964 & 0.969\\
            \cline{1-14}
        \end{tabular*}}\vspace{-8pt}
        %\vspace{0.3cm}
        \caption{Benchmarking Results on the three tasks - Image Classification, Object Detection and Video Classification. \vspace{-10pt}}
        \label{tab:results}
\end{table*}
In our experiments we are able to train successive models in as many few-shot domains as we like, scaling up from training on a single instance of each class in a dataset to training on the full dataset in a single experiment. The ability to run a multitude of $n$-shot settings in a single experiment as opposed to having a user run each experiment individually is both task and algorithm agnostic. This functionality provides a benefit in both efficiency and scalability to the user as it allows them to have a single run that encapsulates the full breadth of few-shot experiments that they would want to run.

Table~\ref{tab:results} provides an overview of the results across the different tasks, algorithms, and datasets for different incremental label budgets ranging from 1-shot to the full size of the training set. Using our framework we show that it is possible to train effectively in the few-shot setting on a number of different datasets across multiple domains and tasks. In total, we train $8$ networks on $6$ different datasets. In each of our few-shot experiments we show that the model scores at least $79\%$ of the accuracy that is achieved on the full dataset with certain domains showing even more progress in the few-shot setting. 
For the Video Classification task, for example, the models trained on $10$ samples from each class scores 98\% of the accuracy that is achieved on the full dataset. This can be attested to the strength of the algorithms and simplicity of the datasets, but it showcases the high level of accuracy that can be achieved even through training on a small number of examples. Figure~\ref{fig:vid_cls_qual_results} contains some sample results for the 1-shot and 4-shot settings using the TimeSformer~\cite{bertasius2021spacetime} method on the UCF101~\cite{soomro2012ucf101} dataset. This example shows that our framework can also handle a custom $n$-shot (4-shot) setting apart from the 1,2,5 and 10-shot results discussed in Table~\ref{tab:results}.

Our object detection models perform quite well in the few-shot setting. DETReg~\cite{bar2022detreg} and CutLER~\cite{wang2023cut}  are able to achieve a mAP of nearly 0.5 on the PoolCar~\cite{poolcar}  dataset with only 2 labels per class as shown in \Cref{tab:results}. Qualitative results for the CutLER pre-training algorithm on PoolCar and xView~\cite{lam2018xview} in the 1-shot and 5-shot settings can be seen in \Cref{fig:obj_det_qual_results}. Our model is able to detect most of the objects on both datasets with very few training labels. The model does particularly well with the ``Small Car" and ``Building" detections, which is expected as these are the two most common classes in xView. The model has some false positives on what appear to be shipping containers, which may in fact be missing ground truth (i.e. not actually false positives). 

We also showcase our use of domain adaptation for the Image Classification problem. In Table~\ref{tab:results}, we show benchmarks on MetaBaseline~\cite{chen2021meta}, MME~\cite{saito2019semi} and PACMAC~\cite{prabhu2022adapting} methods for the Domain-Net~\cite{peng2019moment} Clipart $\xrightarrow{}$ Sketch, Office-Home~\cite{venkateswara2017deep} ClipArt $\xrightarrow{}$ Art and the Office-31~\cite{saenko2010adapting} Webcam $\xrightarrow{}$ DSLR domain-adapt scenarios. 
Figure~\ref{fig:img_cls_qual_results} shows sample images and predictions for the $1$-shot and $5$-shot image classification tasks using the PACMAC~\cite{prabhu2022adapting} algorithm. Figure~\ref{fig:img_cls_qual_results}(A) shows the top-1 predictions comparison between the $1$-shot and $5$-shot tasks for the same image sample from the DomainNet Sketch dataset. Figure~\ref{fig:img_cls_qual_results}(B) shows a similar comparison for the Office-Home Art dataset.

For the DomainNet~\cite{peng2019moment} dataset, PACMAC~\cite{prabhu2022adapting} has the best accuracy with the accuracy score ranging between $48.8\%$ to $56.6\%$ for $1$-shot through $10$-shot tasks. PACMAC also achieves perfect accuracy on the Office31~\cite{saenko2010adapting} dataset. 
However, for the Office-Home ~\cite{venkateswara2017deep} dataset, MME~\cite{saito2019semi} outperforms the other algorithms on the different $n$-shot tasks. Figure \ref{fig:img_cls_qual_results}(A) shows some interesting misclassified results in $1$-shot that are corrected in the $5$-shot task. Out of those examples, the ``Beach" prediction in the $1$-shot task is reasonable since the model would have focused on the water and land elements in the image. Another interesting example in Figure \ref{fig:img_cls_qual_results}(B) is the one that is misclassified in both $1$ and $5$-shot tasks - ``Candles" - where both tasks focus on the illumination in the silhouette image.
\section{Conclusions and Lessons Learned} \label{sec:conclusion}
\textbf{Customizability}. 
The Hydra config protocol provided the basis for customizing the framework across different tasks, datasets and algorithms with the help of fine-grained parameters. For example, for the zero-shot image classification task, we were able to do a straightforward customization of the original image classification task to add a flag to identify if the task is a zero-shot task or not. The framework is thus built with a level of modularity that is scalable to adding new computer vision tasks. When we consider datasets, our framework has the capability to include multiple domain adaptation stages with each stage having an on-the-fly specification of the $k$-shot task, where $k$ is specified as part of the seed budgets for a particular stage. In addition, there is flexibility in terms of the label budgets to extend the framework from a few-shot system to a many-shot classification/detection system.

\textbf{Usability}.
An important aspect to any system is that it behaves the same way for each user. In the context of a training system, this means that all of the packages and tools that the system utilizes must be the same across all devices it is running on. The LEARN system accomplishes this by including a general requirements file specifying package and tool versions as well as specific package files for each of the algorithms that require additional dependencies. This delineation allows an end user to create either a single reproducible environment to run multiple algorithms or to create algorithm specific environments. Either way this guarantees the consistent training and use of the LEARN system through consistent dependency specifications.

%dataset management - automatically matches tareget dataset to source dataset with the most overlapping categories

\section{Future Work} \label{sec:future_work}
The proposed LEARN framework has numerous ways of extending capabilities with respect to the domain-adapt few-shot learning problem. One future feature priority would be the capability to setup thousands of randomly initialized episodic $N$-way tasks where $N$ is the number of classes. In terms of experiment reliability, it is hard to keep track of all the customizable parameters while setting up the Hydra command and can lead to setting up wrong parameter values or sometimes overriding the same parameter twice. To help with this, building a simple GUI for setting parameters would be very a useful extension for the framework.
\section{Acknowledgement} \label{sec:acknowledgement}
This material is based on research sponsored by DARPA under agreement number FA8750‐19‐1‐0504. The U.S. Government is authorized to reproduce and distribute reprints for Governmental purposes notwithstanding any copyright notation thereon.

{
    \small
    \bibliographystyle{ieeenat_fullname}
    \bibliography{main}
}


\end{document}